\documentclass[letterpaper,journal]{IEEEtran}
\usepackage{amsmath,amsfonts}
\usepackage{algorithmic}
\usepackage{algorithm}
\usepackage{array}
\usepackage{booktabs}
\usepackage[caption=false,font=normalsize,labelfont=sf,textfont=sf]{subfig}
\usepackage{textcomp}
\usepackage{stfloats}
\usepackage{url}
\usepackage{verbatim}
\usepackage{graphicx}
\usepackage{cite}
\usepackage[utf8]{inputenc}
\usepackage{amssymb,latexsym}
\usepackage{tikz}
\usetikzlibrary{arrows.meta,backgrounds,fit,positioning}
\begin{document}

\title{Visual Place Recognition Using Rate-Encoded Spiking Neural Networks with Discrete STDP Learning}

\author{Altzi Tsanko, Oikonomou Katerina Maria and Antonios Gasteratos~\IEEEmembership{Senior Member, IEEE}%
\thanks{A. Tsanko, K. Oikonomou, and A. Gasteratos are with the Department of Production and Management Engineering, Democritus University of Thrace, Xanthi 67100, Greece, e-mails: \{altztsan, aioikono, agaster\}@pme.duth.gr.}%
}

\markboth{IEEE Robotics and Automation Letters}%
{Tsanko \MakeLowercase{\textit{et al.}}: Discrete State Isolation in STDP-Based SNN-VPR}

\maketitle

\begin{abstract}
Spiking Neural Networks (SNNs) trained through unsupervised Spike-Timing-Dependent Plasticity (STDP) have been explored as solutions to visual loop closure problems, driven by the prospect of efficient on-device inference on neuromorphic devices. State-of-the-art STDP-based models deliver high classification accuracy but fail to reach the high Recall at 100\% Precision (R@100P) needed for reliable autonomous navigation. We present a discrete, tensor-native implementation of the STDP-based SNN-VPR pipeline using PyTorch with snnTorch and evaluate it on a 100-place Nordland dataset using 15 independently-trained networks. The contribution of three decisions in the implementation is investigated. First, we show how to perform neuron assignment with a closed-form, deterministic tensor pipeline and show that it provides significantly higher R@100P than a standard argmax procedure. However, some of this gain comes from implementation differences compared to prior continuous-time models, which we measure independently. Second, ablation in isolation shows that state reset after each query helps improve R@100P regardless of the way neurons are assigned. Third, velocity-compensated sliding window aggregation over k consecutive frames reaches R@100P = $100.00\% \pm 0.00\%$ at k = 5 for constant-velocity traversal and an additional 0.20 ms latency. Taken together, these findings show the impact of inference stage design decisions in STDP-based SNN-VPR on recall precision, although the separate contribution of each mechanism and implementation differences is only partially disentangled and needs further examination.
\end{abstract}


\section{Introduction}

\IEEEPARstart{V}{ISUAL} Place Recognition (VPR) serves as one of the essential components of SLAM~\cite{lowry2016survey}, allowing the robot to recognize a place it was seen before using only visual data. The primary condition that should be satisfied by VPR for loop closure is not just high classification accuracy, but high retrieval precision, since confident but wrong place match will introduce geometrical errors to robot's map, and therefore will compromise further localization. This is quantified by Recall at 100\% Precision (R@100P), or the amount of correct matches that can be made with the condition that zero false positive places can be used. Obtaining high R@100P proves to be a much harder task compared to classification accuracy, especially under the perceptual aliasing due to cross-seasonal and cross-illumination conditions.

SNNs provide an attractive platform for on-device applications such as VPR.\@ As the third generation of neural networks~\cite{maass1997networks}, SNNs operate on sparse, event-driven spikes and can be trained using STDP, a biologically motivated unsupervised learning algorithm~\cite{bi1998synaptic} avoiding the need for computationally expensive backpropagation. The applicability of this approach proved to go far beyond classification tasks. Spiking actor networks have been used to perform closed-loop robotic arm control~\cite{katerina1}, spiking feature-discrimination architectures have been developed to improve multimodal audio-visual fusion~\cite{katerina2}, three-dimensional spiking convolutional networks have been utilized for in-cabin activity recognition within autonomous vehicles~\cite{katerina3}. However, STDP-based VPR approaches continue to demonstrate substantial gap between classification accuracy and retrieval precision.

This gap is currently not fully understood. While previous works~\cite{Hussaini2022_1,Hussaini2025} demonstrated that going beyond argmax assignment led to tangible improvements in R@100P, there are three potential confounds in separating the effect of the strategy. First, the issue of ambiguity, the proposed assignment strategies were formulated conceptually without giving deterministic formulas, making it impossible to distinguish between the gain coming from the strategy itself and the implementation peculiarities. Second, the problem of conflating simulations, continuous-time ODE solvers like Brian2 cannot guarantee the isolation of neuronal state across independent query presentations. Thus, how much does the temporal leakage influence the reported R@100P?\@ Third, the variance issue, no evaluation of the current approaches reports multiple-runs estimates, making it impossible to distinguish the methodology effect from the seed-to-seed variance of unsupervised STDP training.

The confounds are overcome in this study through a controlled ablation experiment. The primary approach involves a tensor-native ablation implementation using snnTorch in PyTorch, enabling independent toggling of each confound and explicit assignment strategies via tensor operations. The methodology is precisely specified through formalization, state isolation using a binary criterion, and reporting of all outcomes as means and standard deviations across 15 independently trained networks. 

It should be stressed that realizing the energy-efficiency benefit of SNNs necessitates running them on specialized substrate such as Intel Loihi~\cite{davies2018loihi} or BrainScaleS~\cite{schemmel2010wafer}. Energy efficiency of SNNs in GPU simulation is no better and could even be worse compared to ANN inference due to simulation windows needed for SNN simulation. Thus, this paper concerns itself only with precision gains in algorithms within the discrete GPU-native framework. The contribution of this letter include:

\begin{itemize}

\item \textbf{Probabilistic Assignment Formalization as the Primary Precision Driver.} Reformulation of probabilistic assignment and weighted assignment is made into a deterministic tensor pipeline formulation complete with participation regularization, response magnitude normalization, missing spike penalty, and L1-normalized probability output, without referring to any particular ODE solver. Using fifteen independently-trained networks, this tensor pipeline provides a mean R@100P of $77.93\% \pm 3.97\%$ on Nordland in comparison to the baseline assignment which obtains 44.13\%, whereas classification accuracy does not change much ($\approx 94\%$).

\item \textbf{State Isolation as a Contributing Factor.}The within-
system isolation experiment, where architecture, training procedure, and assignment methodology remain unchanged, shows that nullification of membrane potentials and synaptic traces before each query trial increases a bit performance of AUC-PR and positively impacts R@100P.

\item \textbf{Sequence Aggregation as a Sufficient Condition for High-Precision Retrieval.} Summing the probabilities for each frame across the velocity-adjusted sliding window of k frames yields an improvement in the average R@100P from $77.93\% \pm 3.97\%$ when k = 1 to $100.00\% \pm 0.00\%$ when k = 5 at the cost of a further latency of 0.20 ms. This result holds even in the case where the velocity estimate may be wrong by $\pm10\%$.

\end{itemize}

The rest of the letter proceeds in the following way. Related work is discussed in Section~II.\@ Section~III describes the implementation of our solution. The experiment setting is described in Section~IV.\@ Results are presented in Section~V. Section~VI concludes with an agenda for future research.

\section{Related Work}

\subsection{Unsupervised SNN Training}

The first major issue related to the use of SNNs is the nondifferentiable nature of the spiking activation function called the ``dead neuron problem''~\cite{Eshraghian2021,Yamazaki2022}. ANN-to-SNN conversion and BPTT-based training in conjunction with the spiking function show good results, yet the supervised learning signal makes these techniques not applicable to an unsupervised on-device regime that we consider~\cite{Nunes2022,Aribe2025}. The only biological unsupervised learning rule at hand is STDP, and even its GPU-based stochastic implementations remain quite robust with respect to quantization up to 2-bit fixed-point precision~\cite{She2019,Nunes2022}. However, the STDP-trained network has a smaller retrieval precision compared to the supervised counterpart~\cite{Aribe2025}.

\subsection{STDP-Based VPR and Neuronal Assignment}

The earliest bio-inspired VPR systems have explored downsampling in combination with lightweight SeqSLAM techniques using event-camera~\cite{Milford2015}, while other systems like FlyNet have attempted to address significant day-night variations with sparse projections based on insects with less than 75,000 parameters~\cite{Chancan2019}. Most relevant literature is found in~\cite{Hussaini2022_1,Hussaini2022_2,Hussaini2025}, where authors present weighted neuronal assignment, leveraging the full response distribution of a neuron rather than a single peak response as an approach to reduce ambiguity of neurons~\cite{Hussaini2022_1}. Additionally, a probabilistic extension of the idea was presented, where population spike responses are transformed into a normalized confidence distribution~\cite{Hussaini2025}. A related scalable extension focused on ensembling spiking neural networks (SNNs) in regional fashion using Nordland and Oxford RobotCar datasets~\cite{Hussaini2022_2}. There are two issues left unresolved, first, an assignment strategy was only considered conceptually rather than being proposed as closed-form tensor computation. Second, the simulation of a continuous ODE may not isolate the neuronal state for every individual query presentation, hence the effect of time-spillover of the process on performance results cannot be untangled from the assignment procedure. Another bio-inspired direction, VPRTempo, uses supervised delta-rule learning to substitute rate coding by temporal coding, providing an inference greater than 1900 Hz on GPU hardware~\cite{Hines2024}.

\subsection{Sequence-Based Place Recognition}

SeqSLAM and its derivatives improve accuracy through short sequence matching by utilizing the robot’s continuous movement and overcoming any ambiguities in single frame matches~\cite{Milford2015}. Although probability outputs are incorporated into STDP-based SNN-VPR~\cite{Hussaini2025}, the incorporation of both sequence level aggregation and confidence at the population level have never been attempted in previous works. This study makes this innovation in the field for the first time in the STDP-based SNN-VPR context. A different path of research is concerned with VPR from asynchronous event streams using deep spiking encoders with contrastive or residual learning~\cite{Keime2026,Liu2026}.

\subsection{Positioning}

Prior studies~\cite{Hussaini2022_1} proved the significance of the assignment strategy in controlling R@100P in STDP based SNN-VPR and devised weighted and probability based assignment strategies, which form the foundation of this paper. Subsequent researches~\cite{Hussaini2022_2,Hussaini2025} focused on scalability through use of modular ensemble architecture, showed high sensitivity of SNN to sequence matching at R@1 and offered an indicator for sequence matching efficiency. But in all of these studies, argmax assignment was used consistently and no investigation on the behavior of probability based output at R@100P with sequence aggregation was conducted. This leaves three open issues unaddressed. The first concerns the extent to which the assignment strategy contributes to the improvement of R@100P, independent of errors introduced by continuous ODE simulation. The second concerns the seed-to-seed variance among independently trained networks, as opposed to ensembles. The third is whether velocity-compensated probability sequence aggregation can achieve perfect R@100P, and if so, within what operational window. This paper addresses each of these issues in turn.

\section{Methodology}\label{sec:methodology}

To enable controlled analysis of the design choices impacting the retrieval accuracy, the STDP-driven VPR algorithm is implemented as a discrete tensor-based algorithm using PyTorch via snnTorch\footnote{Available at: \protect\url{https://github.com/AltziTS007/torchVPRSNN}.}. With the ability to consider each design decision, assignment policy, isolation of state variables, and temporal summation as individual, separately controllable components, the implementation enables controlled dissection not possible with continuous-time ODE implementations. This framework is based on biological methods outlined in~\cite{Hussaini2022_1}, which are translated into batched step functions eliminating ODE error and enabling GPU optimization while retaining strict unsupervised on-device learning without backpropagation.

\subsection{Image Preprocessing and Spike Encoding}

In the presence of cross-seasonal illumination variations typical of
benchmarks such as the Nordland dataset, severe perceptual aliasing
occurs.
To mitigate this, we apply a patch-based preprocessing pipeline before
encoding.

\subsubsection{Patch Normalization}

Images are converted to greyscale, downsampled to $28\times28$, and
partitioned into non-overlapping $7\times7$ patches.
Localised Min-Max normalization is applied independently to each
patch~$P_k$:

\begin{equation}
\hat{P}_k = \frac{P_k - \min(P_k)}{\max(P_k) - \min(P_k)}
\label{eq:patch-minmax-normalization}
\end{equation}

This inductive bias forces the network to extract high-frequency
structural edges rather than absolute scene luminance, improving
robustness to day-night cycles.

\subsubsection{Poisson Rate Encoding and State Isolation}

The normalised pixel intensities~$\hat{p}_i$ are converted into
discrete spike trains~$s_i[t] \in \{0,1\}$ over a fixed simulation
window ($T=150$ or $T=300$) via a Poisson process where
$P(s_i[t]=1) \propto \hat{p}_i$.

Our discrete framework enforces \textbf{explicit state isolation} by zeroing all membrane potentials and synaptic traces at~$t=0$ for every new query image.
This eliminates the temporal spill-over inherent to continuous ODE
simulations, where decaying traces from previous images may corrupt
the responses to subsequent queries.

\subsection{Discrete Leaky Integrate-and-Fire (LIF) Dynamics}

The architecture comprises a feedforward input layer with
$N_{in} = 784$ inputs and one excitatory layer with~$N_{exc}$ discrete
Leaky Integrate-and-Fire (LIF) neurons.
The membrane potential~$U_j[t]$ of neuron~$j$ evolves iteratively:

\begin{equation}
U_j[t] = \beta U_j[t-1] + I_j[t]
\label{eq:lif-membrane-update}
\end{equation}

Here, $\beta$ is the membrane decay factor, and
$I_j[t] = \sum_{i=1}^{N_{in}} w_{ij} s_i[t]$ is the afferent synaptic
current from input spikes.
A neuron fires $S_j[t]=1$ when its membrane potential crosses an
adaptive threshold~$\theta_j$, after which the potential is reset.

\subsection{Lateral Inhibition via Hard Winner-Take-All (WTA)}

To enforce a sparse, competitive representation, we mathematically
formalize the Hard WTA mechanism over the active set.
Let $\mathcal{A}[t]$ represent candidate neurons whose afferent current
exceeds their adaptive threshold at time~$t$:

\begin{equation}
\mathcal{A}[t] = \bigl\{ k \in \{1,\dots,N_{exc}\} \mid I_k[t] > \theta_k \bigr\}
\label{eq:active-neuron-set}
\end{equation}

A neuron~$j$ fires if and only if it possesses the maximum current
among valid candidates:

\begin{equation}
S_j[t] = \begin{cases}
1 & \text{if } j \in \mathcal{A}[t] \text{ and } I_j[t] = \max_{k \in \mathcal{A}[t]}(I_k[t]) \\
0 & \text{otherwise}
\end{cases}
\label{eq:hard-wta-spike-rule}
\end{equation}

If $\mathcal{A}[t] = \emptyset$, the network remains quiescent.

\subsection{Unsupervised Learning Mechanisms}

\subsubsection{Discrete Weight-Dependent STDP}

We replace continuous, event-driven ODE STDP traces with batch-based
discrete updates.
Pre-synaptic ($x_i$) and post-synaptic ($y_j$) traces decay
iteratively:

\begin{equation}
x_i[t] = x_i[t-1] \!\left(1 - \frac{1}{\tau_{pre}}\right) + s_i[t]
\label{eq:pre-trace}
\end{equation}
\begin{equation}
y_j[t] = y_j[t-1] \!\left(1 - \frac{1}{\tau_{post}}\right) + S_j[t]
\label{eq:post-trace}
\end{equation}

The weight update~$\Delta w_{ij}$ at timestep~$t$ incorporates a soft
upper bound ($w_{\max}$) to prevent runaway growth:

\begin{equation}
\Delta w_{ij} = A_+(w_{\max} - w_{ij})\,x_i[t]\,S_j[t]
              - A_-\,w_{ij}\,y_j[t]\,s_i[t]
\label{eq:stdp-weight-update}
\end{equation}

Following batched STDP updates, $L_2$ normalization is applied
column-wise across the weight matrix:
$w_{ij} \leftarrow w_{ij}/\!\sqrt{\sum_k w_{kj}^2}$.

\subsubsection{Global Homeostatic Stability}

To ensure cohesive place-to-neuron mappings, our framework implements
\textbf{global batch-level homeostasis}.
The threshold~$\theta_j$ is adjusted using the empirical mean firing
rate $\bar{r}_j = \frac{1}{T}\sum_t S_j[t]$ evaluated over the entire
simulation window:

\begin{equation}
\theta_j \leftarrow \theta_j + \eta_\theta\,(\bar{r}_j - r_{target})
\label{eq:homeostatic-threshold-update}
\end{equation}

Where $\eta_\theta$ is the adaptation rate and $r_{target}$ is the
desired sparsity.
Unlike local spike-driven adaptation, batch-level homeostasis prevents
rapid refractoriness from fragmenting the representation of a single
physical location traversed at low speed.

\subsection{Inference and Neuronal Assignments}

Because STDP is purely unsupervised, a calibration phase assigns
neurons to geographic locations.
The training dataset is re-presented with plasticity disabled.
Let $S^R \in \mathbb{N}^{N_{exc} \times N_{classes}}$ represent the
training spike count matrix, where $S^R_{il}$ is the total spikes
emitted by neuron~$i$ for place~$l$.

\subsubsection{Standard Assignment (Baseline)}

A neuron~$i$ is mapped exclusively to the class~$l^*$ that elicits its
maximum average spike count: $l^* = \arg\max_l S^R_{il}$.
The query score for class~$l$ is the sum of spikes~$S^Q_i$ for all
neurons assigned to~$l$.

\subsubsection{Weighted Assignment}

Weighted assignment utilises the full response distribution,
inherently down-weighting ambiguous, non-selective neurons.

\textbf{a. Regularization by Involvement}: let
  $\omega_i = \sum_l \mathbb{I}(S^R_{il} > 0)$ be the number of
  places neuron~$i$ fired for.
  \begin{equation}
  S'_{il} = \begin{cases}
    S^R_{il} \times \dfrac{1}{\omega_i} & \text{if } \omega_i > \gamma \cdot N_{classes} \\[4pt]
    S^R_{il} & \text{otherwise}
  \end{cases}
  \label{eq:involvement-regularization}
  \end{equation}
  Here, $\gamma \in (0,1)$ is the involvement threshold; neurons whose
  responses span more than a $\gamma$ fraction of all classes are
  considered non-selective and penalised by $1/\omega_i$.

\textbf{b. Response Strength Normalization}:
  \begin{equation}
  S''_{il} = S'_{il} \times \frac{S^R_{il}}{\sum_m S^R_{im}}
  \label{eq:response-strength-normalization}
  \end{equation}

\textbf{c. Missing Spikes Penalty}:
  \begin{equation}
  \mathrm{Score}_l =
    \!\left(\sum_i S^Q_i \cdot S''_{il}\right)
    \times
    \frac{\displaystyle\sum_{m \in \mathrm{Active}} S^R_{ml}}
         {\displaystyle\sum_{m} S^R_{ml}}
  \label{eq:missing-spikes-penalty-score}
  \end{equation}
  where $\mathrm{Active}$ is the set of neurons that fired during the query.

\subsubsection{Probability-Based Assignment}

This strategy converts raw weighted scores~$\mathrm{Score}_l$ into a
probability distribution~$P(l \mid Q)$ for each query, providing a
threshold-independent confidence metric.
First, query-specific Min-Max normalization bounds the responses:

\begin{equation}
\widehat{\mathrm{Score}}_l =
  \frac{\mathrm{Score}_l - \min_m(\mathrm{Score}_m)}
       {\max_m(\mathrm{Score}_m) - \min_m(\mathrm{Score}_m)}
\label{eq:score-minmax-normalization}
\end{equation}

Then L1-normalization converts bounded scores into a valid probability
distribution:

\begin{equation}
P(l \mid Q) = \frac{\widehat{\mathrm{Score}}_l}{\sum_{m} \widehat{\mathrm{Score}}_m}
\label{eq:probability-assignment}
\end{equation}

\section{Experimental Setup and Implementation}\label{sec:experimental-setup}

The data sets, network architecture, training procedure, and hardware are specified. We reiterate that the experimental results in this paper have been generated using a proof-of-concept setting in which 100 place benchmarks sequences, constant velocity traversals, and GPU simulations have been used. The specific choice of the experimental settings was made to facilitate a clean factorial ablation of design choices.

\subsection{Datasets and Preprocessing}

Robustness against seasonal and illumination changes is evaluated on two VPR datasets~\cite{Hussaini2022_1,Hussaini2022_2,Hines2024}. In the Nordland dataset, tunnels, stationary parts, and slow-motion periods are ignored for frame alignment, sampling takes place at 100 meters distance apart from the first quartile where spring and autumn passes serve as reference, and summer pass is considered as query. In the urban Oxford RobotCar dataset, samples are collected at every 10 meters interval where sunny and rainy passes act as reference and dusk pass is the query. All images are preprocessed as described in Section~III-A, with $T = 150$ time-steps for Oxford and $T = 300$ time-steps for the Nordland dataset.

\subsection{Network Architecture and Implementation Details}

\begin{figure*}[!t]
    \centering
    \resizebox{0.8\textwidth}{!}{%
    \begin{tikzpicture}[
        node distance=1.2cm and 0.85cm,
        process/.style={rectangle, draw=blue!60, fill=blue!7, rounded corners=3pt, minimum height=1cm, text width=2.75cm, align=center, font=\small},
        endpoint/.style={process, draw=green!55!black, fill=green!10, font=\small\bfseries},
        training/.style={process, draw=orange!85!black, fill=orange!12, dashed, text width=2.5cm, font=\footnotesize},
        calib/.style={process, draw=gray!60!black, fill=gray!15, dashed, text width=2.6cm, font=\footnotesize},
        group/.style={rectangle, draw=gray!60, fill=gray!7, rounded corners=5pt, dashed, inner xsep=0.35cm, inner ysep=0.30cm},
        flow/.style={-{Latex[length=2.4mm]}, thick, draw=blue!60},
        trainflow/.style={-{Latex[length=2.4mm]}, thick, dashed, draw=orange!85!black},
        calibflow/.style={-{Latex[length=2.4mm]}, thick, dashed, draw=gray!60!black}
        ]

        \node[endpoint] (input) {Raw Image};

        \node[process, right=of input] (pre1) {Greyscale \\ \& Downsampling};
        \node[process, right=of pre1] (pre2) {Patch \\ Normalization};
        \node[process, right=of pre2] (pre3) {Spike Encoding \\ (Poisson Rate)};

        \node[process, below=3.05cm of pre3] (snn1) {State Reset \\ (per query, $t{=}0$)};
        \node[process, below=3.05cm of pre2] (snn2) {LIF Neuron \\ Processing};
        \node[process, below=3.05cm of pre1] (snn3) {Winner-Take-All \\ (WTA)};

        \node[training, above=0.55cm of snn2] (train1) {STDP \\ Learning};
        \node[training, above=0.55cm of snn3] (train2) {Adaptive \\ Homeostasis};

        \node[process, below=3.05cm of snn3] (out1) {Spike Count \\ Readout};
        \node[process, below=3.05cm of snn2] (out2) {Neuronal Assignment \\ (W+Prob)};
        \node[process, below=3.05cm of snn1] (out3) {Sequence \\ Aggregation};

        \node[calib, above=0.55cm of out2] (calib) {Calibration \\ (offline, frozen weights)};

        \node[endpoint, right=of out3] (output) {Final Place \\ Prediction};

        \begin{scope}[on background layer]
            \node[group, fit=(pre1) (pre2) (pre3), label={[font=\bfseries]below:Preprocessing \& Encoding}] {};
            \node[group, fit=(snn1) (snn2) (snn3) (train1) (train2), label={[font=\bfseries]below:SNN Core}] {};
            \node[group, fit=(out1) (out2) (out3) (calib), label={[font=\bfseries]below:Inference \& Readout}] {};
        \end{scope}

        \path[flow] (input) -- (pre1);
        \path[flow] (pre1) -- (pre2);
        \path[flow] (pre2) -- (pre3);

        \path[flow, rounded corners=8pt] (pre3.east) -- ++(0.8,0) |- (snn1.east);

        \path[flow] (snn1) -- (snn2);
        \path[flow] (snn2) -- (snn3);

        \path[flow, rounded corners=8pt] (snn3.west) -- ++(-0.8,0) |- (out1.west);

        \path[flow] (out1) -- (out2);
        \path[flow] (out2) -- (out3);
        \path[flow] (out3) -- (output);

        
        \path[trainflow] (snn3.110) -- (train2.250);
        
        \path[trainflow, rounded corners=4pt] (snn3.70) -- ++(0, 0.3) -| (train1.230);
        
        \path[trainflow, rounded corners=4pt] (train2.310) -- ++(0, -0.3) -| (snn2.110);
        
        \path[trainflow] (train1.290) -- (snn2.70);

        \path[calibflow] (calib.south) -- (out2.north);

    \end{tikzpicture}%
    }
    \caption{Proposed SNN-based Visual Place Recognition architecture. Dashed orange paths indicate processes active exclusively during the training phase (parallel updates for STDP and homeostasis driven by WTA). The dashed grey path indicates the one-time calibration step that establishes the fixed neuron-to-place mapping used by every subsequent query; this mapping is not recomputed at inference. State Reset denotes the explicit zeroing of membrane potentials and synaptic traces once at $t=0$ for each new query image, prior to LIF simulation.}
    \label{fig:snn_architecture}
\end{figure*}
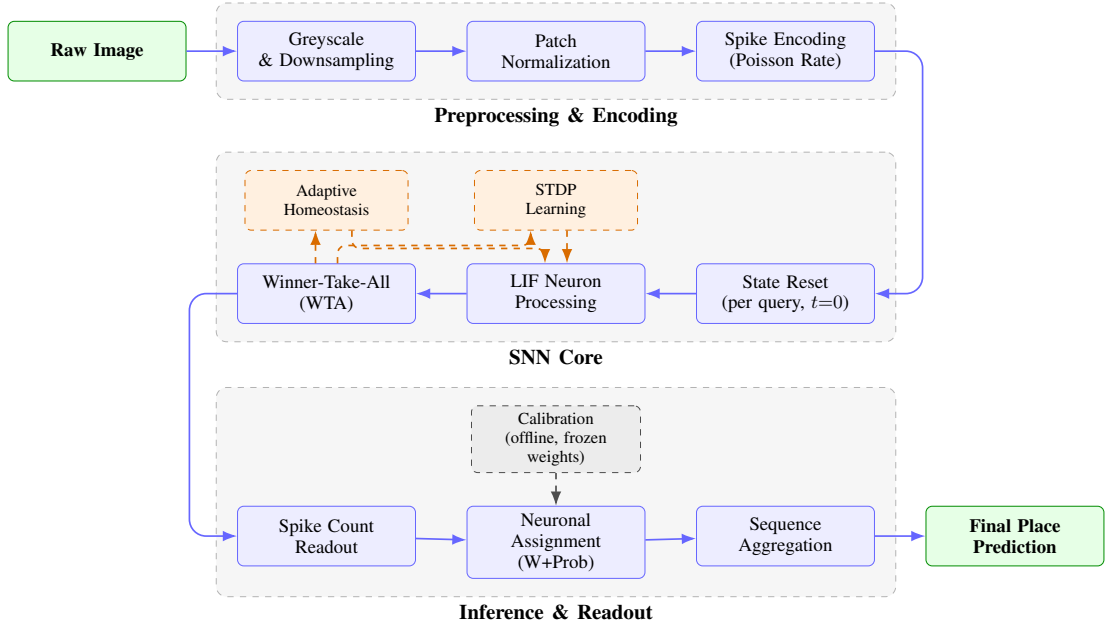

The discrete simulation is implemented natively in PyTorch via snnTorch.
The network comprises an input layer of~$N_{in}=784$ neurons (flattened
image pixels) densely connected to an excitatory population of
$N_{exc}=400$ neurons.
Membrane decay factors are empirically set to~$\beta_{e}=0.95$
(excitatory) and~$\beta_{i}=0.90$ (inhibitory).
Lateral inhibition is enforced via Hard WTA tensor-masking, ensuring
instantaneous suppression of competing neurons.
The initial firing threshold is~$\theta_{init}=1.5$ for excitatory
neurons and fixed at~$1.0$ for inhibitory neurons. The proposed SNN architecture is shown in Fig.~\ref{fig:snn_architecture}.

\subsection{Unsupervised Training Protocol}

Training is purely unsupervised using the discrete weight-dependent
STDP mechanism described in Section~\ref{sec:methodology}, with
learning rates~$A_+=5\times10^{-4}$ (potentiation) and
$A_-=5\times10^{-6}$ (depression), trace decay
constants~$\tau_{pre}=\tau_{post}=15.0$, and synaptic weights clamped
to~$w_{\max}=1.0$.
Global, batch-level homeostatic threshold adaptation maintains a
stable, sparse representation, with target firing
rate~$r_{target}=0.01$ and adaptation rate~$\eta_\theta=0.001$.
Training uses a batch size of~64.

\subsection{Hardware Setup and Evaluation Metrics}

All experiments were performed on a workstation with Ubuntu 22.04 LTS, which includes an Intel Core i7--7700K processor (4.20 GHz), 32 GB of RAM, and an NVIDIA GeForce RTX 3090 graphics card that speeds up tensor computations. Once the unsupervised learning process ends, the plasticity is frozen, and the calibration process associates 400 excitatory neurons with the geographical categories. The results obtained in terms of retrieval performance are assessed using three different neuron assignment methods: Standard Assignment (baseline), Weighted Assignment, and Probability-based Assignment, measured by Accuracy, P@100R, and R@100P.

\section{Results}\label{sec:results}

All results are reported as means and standard deviations across 15 independently trained networks unless otherwise noted.

\subsection{R@100P Across Assignment Strategies}
\label{sec:r100p_improvement}

\begin{figure}[t]
  \centering
  \includegraphics[width=0.48\columnwidth]{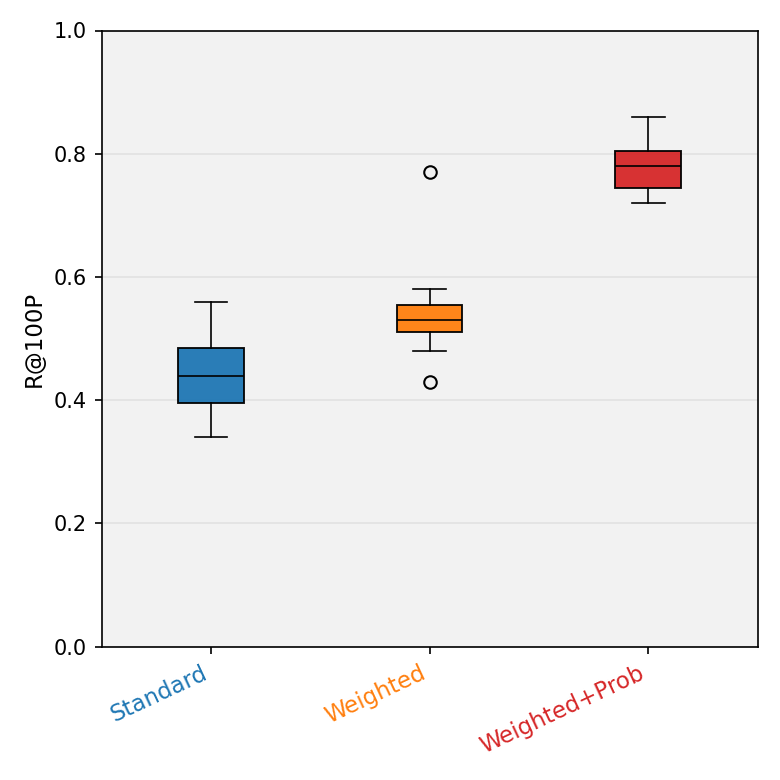}
  \hfill
  \includegraphics[width=0.48\columnwidth]{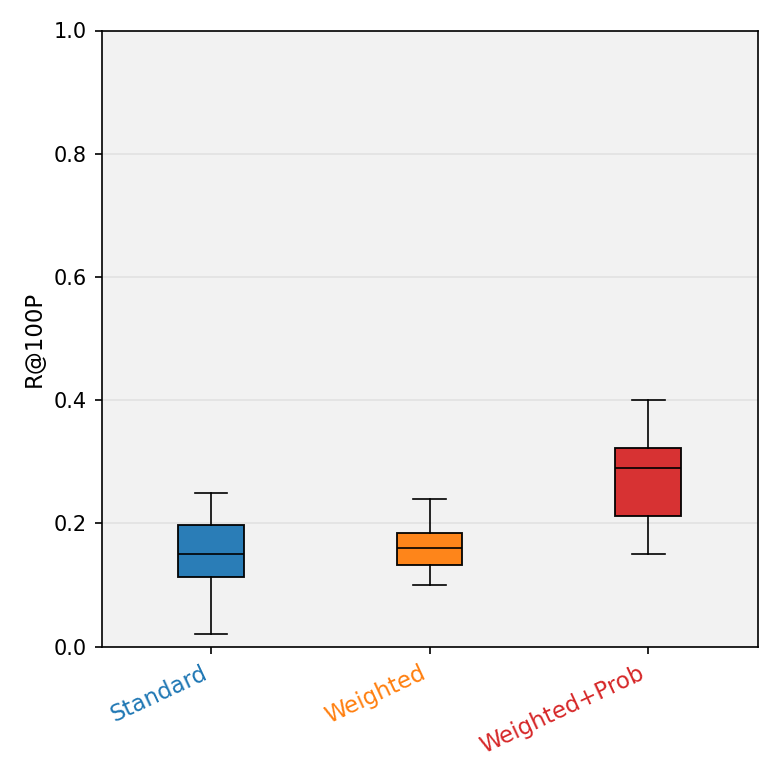}
  \caption{R@100P across assignment strategies on (a)~Nordland (left side) and
    (b)~Oxford RobotCar (right side) with 100-place benchmarks and $N=15$ networks.}
  \label{r100p_comparison}
\end{figure}

In Fig.~\ref{r100p_comparison}, R@100P for all three methods of assignment is shown for Nordland and Oxford with 100 place fields. The method of assignment makes practically no difference to coarse accuracy, which is around 94\% in all cases, but it does make a big difference to the precision of the retrievals. Standard assignment results in a mean R@100P of 44.13\% on Nordland, whereas weighted assignment raises this to 54.13\% by using all the response information from the neuron and punishing ambiguous neurons. Probability-Based Assignment (W+Prob) gives the greatest increase, rising to $77.93\% \pm 3.97\%$, where the upper quartiles are around 86\%.

\subsection{Comparison with Prior Work}
\label{sec:comparison_prior_work}

Table~\ref{tab:results} shows comparisons of our method to the continuous-time ODE SNN-VPR system presented in~\cite{Hussaini2022_1,Hussaini2025}, including NetVLAD~\cite{Arandjelovic2016NetVLAD} and SAD~\cite{Milford2015} for reference as traditional approaches to VPR.\@ Using the Probability-Based Assignment (W+Prob) matching strategy, our discrete state-isolated system achieves an average R@100P of 77.9\% on Nordland, in contrast to 47.5\% of the previous ODE system, a difference of 30.4 percentage points. As explained in Section~V-F, this gap mainly arises due to different codebases and discretization techniques, and not simply state isolation. The Standard and Weighted baselines consistently outperform their continuous-time variants on both datasets. On Oxford RobotCar dataset, the highest performance achieved is 27.1\% R@100P with W+Prob, owing to more viewpoint changes in the urban setting. Sequence aggregation on Oxford and a comparative analysis against other recent methods (Patch-NetVLAD~\cite{Hausler2021PatchNetVLAD}, SeqNet~\cite{Garg2021SeqNet}, AnyLoc~\cite{Keetha2023AnyLoc}) on larger maps is left for future work (Section~VI).

\begin{table*}[!t]
\caption{Mean R@100P Performance Comparison. Continuous-ODE SNN-VPR
  results from~\cite{Hussaini2022_1,Hussaini2025}. NetVLAD and SAD are
  included as contextual references only and are not evaluated under the
  same protocol as the SNN systems. Sequence aggregation results on
  Oxford RobotCar and large-scale evaluations are deferred to future
  work.}
\label{tab:results}
\centering
\scriptsize
\renewcommand{\arraystretch}{1.15}
\setlength{\tabcolsep}{4pt}
\begin{tabular*}{\textwidth}{@{\extracolsep{\fill}}lcccccccc@{}}
\toprule
& \multicolumn{3}{c}{\textbf{Continuous-ODE SNN-VPR~\cite{Hussaini2022_1,Hussaini2025}}}
& \multicolumn{2}{c}{\textbf{Traditional VPR (Reference Only)}}
& \multicolumn{3}{c}{\textbf{snntorchVPR (Ours)}} \\
\textbf{Dataset}
  & \textbf{Standard} & \textbf{Weighted} & \textbf{W+Prob}
  & \textbf{NetVLAD}  & \textbf{SAD}
  & \textbf{Standard} & \textbf{Weighted} & \textbf{W+Prob} \\
\midrule
Nordland        & 12.8\% & 29.7\% & 47.5\% & 15.3\% & 51.7\% & 44.13\% & 54.13\% & 77.93\% \\
Oxford RobotCar &  2.5\% &  8.8\% & 12.0\% & 11.0\% & 11.3\% & 15.2\%  & 16.3\%  & 27.1\%  \\
\bottomrule
\end{tabular*}
\end{table*}

\subsection{Computational Profile}

The tensor-based implementation takes 17.52 ms to perform the entire computation of T = 300 time-steps, consuming about 380 MFLOPS, while the contribution from sequence aggregation is only 0.20~ms. This results in an overall latency of 17.72 ms, achieving throughput of 56.4 Hz, which surpasses the traditional 30 Hz frequency of robot cameras. As stated in Section I, demonstration of the efficiency of our approach in terms of power consumption against that of artificial neural networks requires neuromorphic hardware.

\subsection{Ablation Studies}\label{sec:ablation}

\subsubsection{Core Mechanism Contributions}

As shown in Table~\ref{tab:core-ablation}, the contributions of Hard Winner-Take-All (WTA), global homeostasis, and patch normalization are all necessary for the stable generation of place cells, since their elimination results in nearly total failure of the model. Hard WTA generates competitive representations and does not allow overlapping visual features to fire together. Batch-level homeostatic threshold adaptation counteracts the effect of overstimulation caused by the most dominant neurons and allows the remaining neurons to become active. Patch normalization reduces the luminance variance and highlights the local structural patterns necessary for successful place recognition. 

\begin{table}[t]
\caption{Core Network Component Ablation (100-place Nordland,
  W+Prob assignment).}
\label{tab:core-ablation}
\centering
\scriptsize
\begin{tabular}{lcc}
\toprule
\textbf{Configuration}               & \textbf{Acc.~(\%)} & \textbf{R@100P~(\%)} \\
\midrule
\textbf{Full Architecture}           & \textbf{94.07}     & \textbf{77.93}       \\
w/o Hard WTA                         & $\sim$1.00         & 0.00                 \\
w/o Global Homeostasis               & $\sim$1.00         & 0.00                 \\
w/o Patch Normalization              & $<$5.00            & 0.00                 \\
\bottomrule
\end{tabular}
\end{table}

\subsubsection{Neuronal Assignment Strategies}

According to Table~\ref{tab:assignment-ablation}, the effect of assignment strategy on coarse accuracy is minimal, while its impact on retrieval accuracy is significant. The probability-based strategy (W+Prob) reduces the effect of perceptual aliasing by treating spike numbers as probability distributions and thus reducing the noise of secondary activations, leading to the average R@100P rate of 77.93\%.

\begin{table}[t]
\caption{Assignment strategy ablation on 100-place Nordland (mean across
$N=15$ seeds). W+Prob denotes weighted probabilistic assignment.}
\label{tab:assignment-ablation}
\centering
\scriptsize
\begin{tabular}{lccc}
\toprule
\textbf{Strategy} & \textbf{Acc.} & \textbf{P@100R} & \textbf{R@100P} \\
 & (\%) & (\%) & (\%) \\
\midrule
Standard & 93.40 & 93.40 & 44.13 \\
Weighted & \textbf{94.07} & \textbf{94.07} & 54.13 \\
W+Prob & \textbf{94.07} & \textbf{94.07} & \textbf{77.93} \\
\bottomrule
\end{tabular}
\end{table}

\subsubsection{Scalability Analysis}
\label{sec:scalability}
\begin{figure}[t]
     \centering
     \includegraphics[width=0.8\columnwidth]{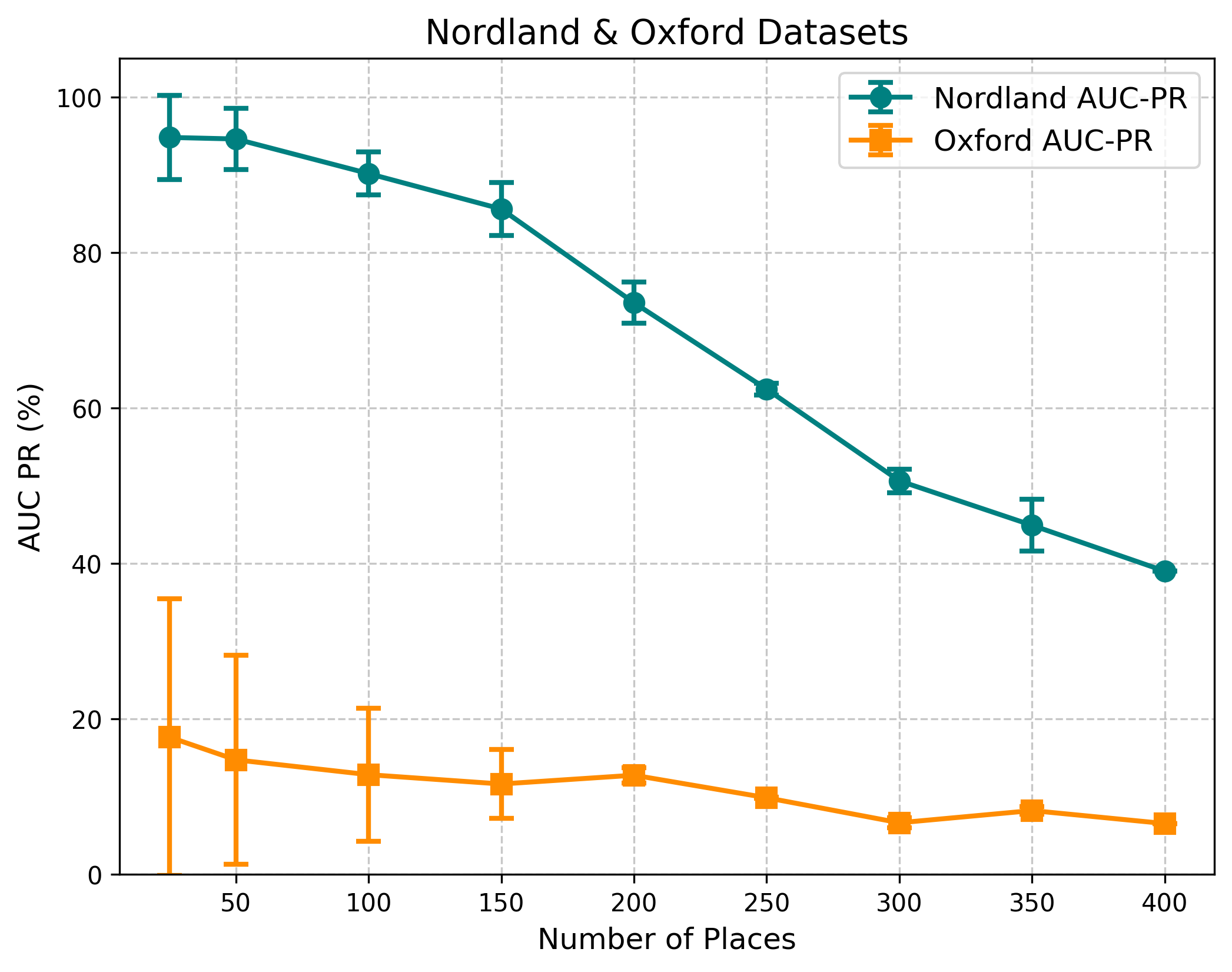}
     \caption{AUC-PR as a function of the number of reference places on Nordland ($T{=}300$) and Oxford RobotCar ($T{=}150$). Error bars denote $\pm 1$ standard deviation across sliding-window segments. The Nordland curve closely reproduces the trend reported in~\cite{Hussaini2022_1}, confirming that the discrete-time formulation faithfully
     reproduces the scalability characteristics of the
     continuous-time ODE baseline.}
     \label{fig:scalability}
 \end{figure}

To evaluate the scalability of the proposed SNN-based VPR system, we examine the effect of the number of reference places $N$ on AUC-PR by using the sliding-window evaluation protocol outlined in~\cite{Hussaini2022_1}. For each $N \in \{25, 50, 100, 150, 200, 250, 300, 350, 400\}$, the dataset is split into overlapping windows of $N$ sequential places, where the network is trained and evaluated independently on each window with corresponding mean and standard deviation of AUC-PR being computed. The Nordland experiments are carried out with $T = 300$ simulation timesteps, while the Oxford RobotCar experiments are done with $T = 150$. Fig.~\ref{fig:scalability} illustrates the results of these experiments. On Nordland, the network obtains $94.83\pm5.41\%$ AUC-PR for $N = 25$ and gracefully degrades to 39.02\% for $N = 400$. On Oxford RobotCar, the performance AUC-PR is substantially worse for all scales considered ($17.66\pm17.77\%$ for $N = 25$ down to 6.54\% for $N = 400$), due to higher visual complexity and perceptual aliasing in the urban environment compared to the railroad corridor of the Nordland dataset. The high variance for small $N$ in the Oxford dataset demonstrates that the performance is highly dependent on the choice of the geographic segment, since some areas of the route are more discriminative visually than others. Overall, there are two major insights obtained from these results: (i) the discrete-time SNN model is comparable with the continuous-time ODE-based model on the Nordland dataset without the need of an ODE solver and is thus suitable for fast GPU-accelerated batch training; and (ii) the non-modular single SNN has fundamental capacity limitations when applied to a challenging urban visual environment, which is the reason for the sequence aggregation method presented in Section~V-E.

\subsubsection{Operational Envelope: Velocity Mismatch and Traversal Discontinuities}
Table~\ref{tab:robustness} depicts deployment constraints of the aggregation scheme ($k = 5$, product rule) on average over 10 trained networks (seeds 16--25). The network operates stably with velocity errors in the range of $\pm10\%$ ($100.00\% \pm 0.00\%$) but fails rapidly at $\pm20\%$ mismatch ($2.92\% \pm 5.90\%$). However, its failures differ depending on the direction of the velocity mismatch, being faster than required does not completely ruin window overlaps ($+20\%$: $5.83\% \pm 7.25\%$), while being slower causes complete failure of the retrieval ($-20\%$: $0.00\%$). Traversal reversals prevent the generation of any proper windows. Frames dropping is successfully overcome by the algorithm: with $10\%$ frame drops R@100P is equal to $100.00\% \pm 0.00\%$, while at $50\%$ frame drops R@100P falls down to $90.00\% \pm 30.00\%$. Thus, high drop ratios can cause some networks to fail to provide enough contiguous windows for aggregation. Route deviations reset windows locally but do not affect retrieval accuracy in other windows ($100.00\% \pm 0.00\%$).

\begin{table}[t]
\caption{Operational envelope of sequence aggregation ($k=5$, product
rule), Nordland, 100 places. Results are averaged over 10 random seeds
(16--25).}
\label{tab:robustness}
\centering
\scriptsize
\begin{tabular}{lcc}
\toprule
\textbf{Condition} & \textbf{Mean R@100P (\%)} & \textbf{Std Dev} \\
\midrule
Baseline ($1.0\times$)              & 100.00 & $\pm$0.00 \\
Velocity mismatch ($+10\%$)         & 100.00 & $\pm$0.00 \\
Velocity mismatch ($-10\%$)         & 100.00 & $\pm$0.00 \\
Velocity mismatch ($+20\%$)         &   5.83 & $\pm$7.25 \\
Velocity mismatch ($-20\%$)         &   0.00 & $\pm$0.00 \\
Velocity mismatch ($+50\%$)         &   0.00 & $\pm$0.00 \\
Velocity mismatch ($-50\%$)         &   0.00 & $\pm$0.00 \\
Reverse traversal                   &   0.00 & $\pm$0.00 \\
10\% dropped frames                 & 100.00 & $\pm$0.00 \\
50\% dropped frames                 &  90.00 & $\pm$30.00 \\
Route deviations                    & 100.00 & $\pm$0.00 \\
\midrule
\multicolumn{3}{c}{\textit{Combined Velocity Mismatches}} \\
\midrule
Velocity mismatch ($\pm10\%$)       & 100.00 & $\pm$0.00 \\
Velocity mismatch ($\pm20\%$)       &   2.92 & $\pm$5.90 \\
Velocity mismatch ($\pm50\%$)       &   0.00 & $\pm$0.00 \\
\bottomrule
\end{tabular}
\end{table}

\subsection{Sequential Aggregation Performance}
\label{sec:seq_agg_performance}

We perform velocity-based sliding window aggregation following the product (log sum) rule by testing 15 networks with $k$ from $\{1, 3, 5, 7, 10, 15\}$ as in Section~V-A. We present the results in Table~\ref{tab:seq-agg}. For $k = 1$, we get the mean R@100P to be $77.93\% \pm 3.97\%$. Going to $k = 3$ brings us an increase to $99.93\% \pm 0.25\%$. For $k \geq 5$, we have $100.00\% \pm 0.00\%$ retrieval for all 15 networks, which holds true until $k = 15$. The greatly decreased standard deviation at $k \geq 3$ in comparison to the rather high standard deviation at $k = 1$ suggests that in addition to raising the mean performance of sequence aggregation helps to reduce the variability of seeds used during unsupervised STDP training, those seeds that provide a low performance at $k = 1$ yield near-perfect retrieval after incorporating temporal context. The overhead of the aggregation at $k = 5$ is 0.20~ms per query.

\begin{table}[ht]
\caption{Sequential Aggregation: Mean R@100P vs.\ Window Size~$k$
  (Nordland, 100~places, constant-velocity traversal, product rule,
  mean~$\pm$~std, $N=15$ networks).}
\label{tab:seq-agg}
\centering
\begin{tabular}{ccc}
\toprule
$k$ & Mean R@100P~(\%) & Std.\ Dev.~(\%) \\
\midrule
1   & 77.93   & 3.97 \\
3   & 99.93   & 0.25 \\
5   & 100.00  & 0.00 \\
7   & 100.00  & 0.00 \\
10  & 100.00  & 0.00 \\
15  & 100.00  & 0.00 \\
\bottomrule
\end{tabular}
\end{table}

\subsection{State Isolation: Within-System Ablation}
\label{sec:state_isolation_ablation}

Turning off the explicit state isolation in our discrete pipeline allows us to observe whether temporal spillover (active membrane potentials and synaptic traces across questions) can impact results while other factors remain equal. Table~\ref{tab:state_isolation} presents results of our experiment on one representative network for the 100-place Nordland benchmark. This is a targeted ablation study on a single run, but multi-seed replication on the entire 15-seed set is planned for the future, as mentioned in Section~VI.\@ Our reported 6-point attribution of R@100P should be viewed cautiously due to the presence of seed-to-seed variability ($\sigma = 3.97\%$ at $k = 1$). Disabling the state isolation results in lower accuracy and AUC-PR for all three assignment strategies. The largest reduction is observed for W+Prob (accuracy decreases from $95\%$ to $90\%$; AUC-PR decreases from $94.42\%$ to $89.21\%$). R@100P is decreased for Standard ($45\%$ to $41\%$) and W+Prob ($71\%$ to $65\%$) assignments. In Weighted condition, we have seen a small increase ($50\%$ to $51\%$), which is most likely due to single-run noise.

Two conclusions can be drawn from this study. First, the 6-point reduction of R@100P within the system for W+Prob is much less significant than 30.4 points gap between systems in Section~V-B. This means that codebase and time steps difference accounts for the most part of that gap, and not the state isolation itself. Second, applying sequence aggregation at $k = 5$ solves the problem of time step spillover completely---with or without isolation, R@100P becomes $100.00\%$ ($N = 96$ sliding windows). It shows that the temporal aggregation makes the algorithm insensitive to both STDP seed variance and state management issues.

\begin{table}[t]
\centering
\caption{Within-System Ablation of Query-State Isolation (Nordland, 100 Places; W+Prob Assignment)}
\label{tab:state_isolation}
\scriptsize
\resizebox{\columnwidth}{!}{%
\begin{tabular}{@{}llcccc@{}}
\toprule
\textbf{Method} & \textbf{State} & \textbf{Acc.~(\%)} & \textbf{P@100R~(\%)} & \textbf{R@100P~(\%)} & \textbf{AUC-PR~(\%)} \\
\midrule
Standard        & Isolated  & 94.00 & 94.00 & 45.00 & 91.47 \\
Standard        & Spillover & 87.00 & 87.00 & 41.00 & 82.49 \\
Weighted        & Isolated  & 95.00 & 95.00 & 50.00 & 93.90 \\
Weighted        & Spillover & 90.00 & 90.00 & 51.00 & 88.05 \\
W+Prob          & Isolated  & 95.00 & 95.00 & 71.00 & 94.42 \\
W+Prob          & Spillover & 90.00 & 90.00 & 65.00 & 89.21 \\
SeqAgg ($k=5$)  & Isolated  & 100.00 & 100.00 & 100.00 & 100.00 \\
SeqAgg ($k=5$)  & Spillover & 100.00 & 100.00 & 100.00 & 100.00 \\
\bottomrule
\end{tabular}%
}
\end{table}

\section{Conclusion}\label{sec:conclusion}

The controlled ablation study on the inference-stage design choices governing retrieval precision of STDP-based SNN-VPR was performed using a discrete tensor-native pipeline in PyTorch with snnTorch. We identified three necessary/sufficient design choices ensuring high-precision retrieval on the used benchmark.

The first design choice, formalization of probabilistic assignment as a closed-form tensor pipeline, is a major source of precision. It increased the mean R@100P from 44.13\% to 77.93\%, while maintaining the coarse accuracy around $94\%$.

The second choice, verified with the help of an in-system ablation study, is that of explicit state isolation, which positively influences retrieval performance regardless of assignment strategy and increases R@100P in Standard and W+Prob strategies. Yet, its isolated contribution of 6 percentage points in the case of W+Prob assignment in one representative network accounts only partially for the 30.4 percentage point gap between our results and those of~\cite{Hussaini2022_1}.

The third, sufficient condition for perfect retrieval on 100-place Nordland benchmark under the assumption of constant-velocity traversal is velocity-compensated sequence aggregation, which results in $100.00\% \pm 0.00\%$ recall at $k = 5$ on all 15 networks without change of any other upstream component. The variance between seeds disappears starting with $k = 3$, suggesting that temporal aggregation compensates for stochasticity during STDP training as well. The operational envelope, stable within $\pm10\%$ velocity error range, with a sharp failure point at $\pm20\%$ defines the practical engineering target for further velocity-robust inference approaches.

Next steps are intended to, validate on neuromorphic hardware (Intel Loihi, BrainScaleS or FPGA), assessing the energy-efficiency of the inference pipeline as compared to ANN inference.
Perform large-scale multi-dataset evaluation on Nordland Sequence Benchmark, Brisbane-Event-VPR, and other datasets with real viewpoint variance. Compare to the state-of-the-art solutions such as PatchNetVLAD~\cite{Hausler2021PatchNetVLAD}, SeqNet~\cite{Garg2021SeqNet}, AnyLoc~\cite{Keetha2023AnyLoc}, VPRTempo~\cite{Hines2024} and foundation-model-based VPR at similar map scale and replicate the multi-seed state-isolation ablation.
Finally extend the velocity tolerance to beyond $\pm10\%$ through adaptive velocity estimation and window probability calculation.

\bibliographystyle{ieeetr}
\bibliography{bibl}

\vfill

\end{document}